\documentclass[11pt]{article}
\usepackage[margin=1in]{geometry}
\usepackage[T1]{fontenc}
\usepackage[utf8]{inputenc}
\usepackage{natbib}
\usepackage{booktabs}
\usepackage{array}
\usepackage{amsmath}
\usepackage{hyperref}
\usepackage{xspace}

\newcommand{\ldv}{LDV-RSI\xspace}

\title{Local-First RSI: Deterministic-Verifier Self-Improvement\\
Loops for LLM Agents on Apple Silicon}

\author{Xihai Wang\thanks{Huawei. Code and raw results:
\url{https://github.com/dahai80/fusion-harness} (loop kernel, PyPI
\texttt{fusion-harness} 0.34.0).} \and Kong Weibu\thanks{Huawei.}}
\date{November 30, 2026}

\begin{document}
\maketitle

\begin{abstract}
Self-improvement loops for LLM agents---propose an edit, evaluate, keep
if better---have only been demonstrated with cloud frontier models as
the optimizer, and graders that are either LLM judges, humans, or
training-time executors in RL regimes. We present \ldv, the first
self-improvement loop operating entirely on consumer Apple Silicon: a
local open-weight proposer (8B) mutates a compaction prompt, a local
open-weight model (4B) executes it, and the keep/rollback gate is a
frozen deterministic verifier---a pure function of output versus
fixture, scored continuously and gated by a paired sign-flip
permutation test. The loop's only cross-round state is a verdict-gated
evolution memory of SIM-shaped records. Across 9 loop runs (3 seeds
$\times$ 3 information arms) on an M5 Max, every run whose gate
accepted $\geq 2$ candidates improved 2.5--3.1$\times$ over the
baseline prompt (0.650 $\rightarrow$ 1.6--2.0 on a 2.0 oracle ceiling,
zero cloud calls), at ${\sim}21$ wall-clock minutes per kept
improvement with the gate itself costing zero model tokens. Three
negative results are as load-bearing as the positive ones: (1) pooled
two-sample gates are systematically conservative on heterogeneous task
sets and stall the loop---pairing fixes it; (2) a weak proposer starves
without verbatim failure evidence, and evolution-memory summaries alone
cannot be shown to add value on top of evidence (all arm means fall
inside the ${\sim}0.3$--0.4 re-score noise floor); (3) across three
gated trials with 11 planted grader-hack decoys, a local 32B LLM judge
kept 8/11 decoys and never selected the genuine improvement, while the
deterministic gate kept 0/11 decoys and the genuine candidate in every
run---at local scale, where the proposer and judge share a capacity
class, selection signals must not share a loss surface with the thing
they grade. We release the loop kernel as a pure-stdlib library and
give a budget accounting for RSI under unified-memory constraints.

\end{abstract}

\section{Introduction}\label{sec:intro}
Recursive self-improvement (RSI) for LLM agents has a proven pattern:
propose an edit to the agent's own harness or prompts, run an
evaluation, keep the edit if it scores better. The pattern works---SICA
raised a coding agent from 17\% to 53\% on SWE-bench
\citep{sica2025}, the Darwin G\"odel Machine from 20\% to 50\%
\citep{dgm2025}---but every published instance makes three assumptions:
a frontier cloud model somewhere in the loop as the optimizer, a grader
that is a benchmark score, an LLM judge, or a training-time executor,
and a GPU-cluster budget. The closest prior verifier design, Absolute
Zero's deterministic code-execution checker \citep{absolutezero2025},
is still an RL-training regime on datacenter hardware, and it verifies
generated tasks rather than the optimizer's own behavioral outputs. No
published loop operates in the on-device constraint regime.

This paper studies that regime. The optimizing agent and the served
model are open-weight peers (4B--14B) sharing unified memory on a
consumer Apple Silicon laptop, served by MLX, with zero cloud calls
anywhere in the loop. The regime inverts the usual threat model: the
proposer is weak enough that noisy selection, not capability, is the
binding constraint---so verifier design and memory discipline matter
more, not less, than at cloud scale.

\paragraph{Contributions.}
\begin{itemize}
\item \textbf{Regime definition.} We formalize the local-first RSI
regime and show empirically what changes when the optimizer is a
peer-capacity local model rather than a frontier oracle (Section
\ref{sec:loop}).
\item \textbf{Deterministic-verifier gate.} A frozen, hidden,
pure-function verifier (token-containment probes over fixtures, scored
continuously, gated by a seeded paired sign-flip permutation test with
noise-band floors) replaces the LLM judge. We plant grader-hacking
decoys in the style of DGM's node-114 incident \citep{dgm2025} and
measure decoy keep-rate: deterministic gate vs.\ a local 32B LLM judge
(E2, Section \ref{sec:results}).
\item \textbf{Evolution memory at local scale.} A verdict-gated,
component-scoped evolution memory (SIM-shaped records: rationale,
predicted and observed outcomes, evaluation context) is the only state
carried across rounds. With N=3 seeds per arm, memory-only
$\approx$ evidence-only $\approx$ memory+evidence: all means fall
inside the ${\sim}0.3$--0.4 re-score noise floor and all far above
baseline---the gate pass rate, not the information source, separates
improving runs from flat ones (E1).
\item \textbf{Budget accounting.} Wall-clock, memory pressure, and cost
per accepted improvement on Apple Silicon, reported against
literature-reported cloud-loop costs (E3): ${\sim}21$ minutes per kept
improvement, gate cost exactly zero model tokens.
\end{itemize}

\section{Related Work}\label{sec:related}
\paragraph{Scaffold/harness evolution.} SICA \citep{sica2025} and the
Darwin G\"odel Machine \citep{dgm2025} establish the propose-eval-keep
pattern for agent-harness code, with cloud frontier models as
optimizers; the Google quartet Harness2/Evo2/RRSI/OO-Harness
\citep{equalcompute2026} adds holdout-attributed deltas across harness
layers; AlphaEvolve scales the pattern to algorithm discovery.
Training-side self-improvement neighbors: RISE \citep{rise2024} teaches
introspection through RL on self-generated traces, and Absolute Zero
\citep{absolutezero2025} derives its own tasks with a deterministic
code-execution verifier---the closest prior use of a non-gameable
grader, but in an RL regime that requires datacenter GPUs and weight
updates. All of the above assume GPU-cluster budgets.

\paragraph{Memory for self-improvement.} CRSI/SIM \citep{crsi2026}
reports +11.33 cross-model transfer from structured evolution records
at cloud scale; Voyager's skill library \citep{voyager2023} and
Reflexion's episodic memory \citep{reflexion2023} are the agent-side
precursors. Our E1 tests whether this replicates when the memory
consumer is a peer-capacity local model, and finds the transfer benefit
does not dominate direct failure evidence.

\paragraph{Prompt optimization.} GEPA \citep{gepa2025}---reflective
prompt mutation against any OpenAI-compatible endpoint, beating GRPO by
+6\% at 35$\times$ fewer rollouts---is the closest optimization recipe;
our proposer is a GEPA-style reflective mutator with the cloud endpoint
replaced by MLX on unified memory. TextGrad \citep{textgrad2024} and
OPRO \citep{opro2023} share the reflect-and-mutate shape with known
variance issues.

\paragraph{Safety / verification.} DGM's node-114 incident
\citep{dgm2025} documented a candidate hacking its own grader;
Winner's-Curse holdout discipline \citep{winnerscurse2026} motivates
our sealed split; equal-compute retest \citep{equalcompute2026}
motivates our budget accounting; AgentPoison \citep{agentpoison2024}
shows memory poisoning attacks that motivate verdict-gated writes. E2
plants DGM-style decoys against both gate types.

\paragraph{On-device inference.} The MLX and llama.cpp-class ecosystems
serve open-weight models on consumer Apple Silicon, but no published
RSI loop operates in this regime---that is the gap this paper fills.
Table \ref{tab:positioning} positions \ldv{} against the closest
systems.

\begin{table}[t]
\centering\footnotesize
\begin{tabular}{>{\raggedright\arraybackslash}p{2.1cm}>{\raggedright\arraybackslash}p{2.3cm}>{\raggedright\arraybackslash}p{2.5cm}>{\raggedright\arraybackslash}p{1.3cm}>{\raggedright\arraybackslash}p{1.9cm}>{\raggedright\arraybackslash}p{1.9cm}>{\raggedright\arraybackslash}p{1.7cm}}
\toprule
System & Optimizer & Grader & Weight training & Hardware & Memory & Transfer \\
\midrule
SICA \citep{sica2025} & cloud frontier & benchmark score & no & datacenter GPU & code archive & n/a \\
DGM \citep{dgm2025} & cloud frontier & SWE-bench tests & no & datacenter GPU & population archive & limited \\
RISE \citep{rise2024} & cloud model, self-generated traces & benchmark score & yes (RL) & datacenter GPU & none across runs & n/a \\
Absolute Zero \citep{absolutezero2025} & cloud frontier, self-play & code execution (deterministic) & yes (RL) & datacenter GPU & none across runs & n/a \\
GEPA \citep{gepa2025} & cloud frontier (any OpenAI-compat.\ endpoint) & task metric + LLM reflection & no & datacenter GPU & per-run candidates & reflection swap shown \\
CRSI/SIM \citep{crsi2026} & cloud frontier & benchmark suite & no & datacenter GPU & SIM records & yes, +11.33 \\
Google quartet \citep{equalcompute2026} & cloud frontier & benchmark + holdout & no & datacenter GPU & partial (Evo2) & across harness layers \\
\textbf{\ldv (ours)} & \textbf{local open-weights (MLX)} & \textbf{deterministic oracle, frozen + hidden} & \textbf{no} & \textbf{consumer laptop} & \textbf{verdict-gated evolution memory} & \textbf{measured E1/E2} \\
\bottomrule
\end{tabular}
\caption{Positioning against the closest self-improvement systems.}
\label{tab:positioning}
\end{table}

Absolute Zero is the closest prior verifier design---its
code-execution checker is deterministic like ours---but it operates in
an RL training regime on datacenter GPUs and verifies generated tasks,
not the optimizer's own behavioral outputs. Every other row assumes a
cloud frontier model somewhere in the loop and an LLM-judge or
benchmark-as-judge. Ours removes all three assumptions at once: the
optimizer is a peer-capacity local model, the grader is a pure function
no proposer output can flatter, and nothing is trained---the whole loop
fits on one laptop.

\section{The Loop: \ldv}\label{sec:loop}
\ldv{} has five components (all pure-stdlib kernel code in
fusion-harness, released on PyPI as version 0.34.0):

\begin{enumerate}
\item \textbf{Proposer} --- local model via an MLX server;
GEPA-style reflective mutation on failure transcripts
\citep{gepa2025}.
\item \textbf{Deterministic verifier} --- frozen and hidden from the
proposer: token-containment probes over fixtures plus a
non-parametric permutation comparator. Never an LLM judge; never in
the optimization loop.
\item \textbf{Evolution memory} --- verdict-gated JSONL records shaped
after SIM \citep{crsi2026}; a summarizer renders the decisive lessons
for the next proposer round; component-scoped and leak-free.
\item \textbf{Regression gate} --- a holdout split never shown to the
optimizer (Winner's-Curse discipline, \citealp{winnerscurse2026});
multi-trial scoring; equal-compute retest
\citep{equalcompute2026}.
\item \textbf{Sandbox + audit} --- candidate edits land as artifacts,
never in-place; archive-don't-overwrite; records carry dataset names
only, never prompt or completion content.
\end{enumerate}

\paragraph{Threat model.} The proposer cannot see the grader; memory
writes are validated like code (a gate failure is a build failure); the
monitor sits outside the optimization loop \citep{monitor2025}.

\section{Experimental Design}\label{sec:design}
All runs execute on one machine (Apple M5 Max, 128\,GB unified memory)
with MLX serving open-weight models: proposer Qwen3-8B, compactor
Qwen3.5-4B-MLX-4bit, judge Qwen3-32B-4bit where E2 needs one. Zero
cloud calls. Tasks are deterministic compaction fixtures with a
pure-function oracle (marker recall + rule recall, ceiling 2.0).

\paragraph{E1 --- does evolution memory compound locally?} The loop
runs with three information arms: A = evolution memory + verbatim
failure evidence, B = evidence only, C = memory only. N=3 seeds per
arm, 10 rounds each. Metrics: final incumbent score on the oracle,
keep-rate, repeated-mistake rate. Hypothesis under test: the SIM
cross-model transfer effect \citep{crsi2026} replicates at local scale.

\paragraph{E2 --- deterministic verifier vs.\ LLM-judge gate.} Same
proposer and task set; the gate is either the deterministic oracle or a
local 32B LLM judge scoring compacted outputs on a 0--10 rubric.
Metrics: true gain on a sealed holdout, grader-hack decoys survived,
gate cost. Hypothesis: deterministic gate equal-or-better true gain,
strictly better anti-hacking, cheaper.

\paragraph{E3 --- budget accounting under unified memory.} Wall-clock,
memory pressure, and tokens per accepted improvement, reported against
cloud-loop baselines as stated in the literature (marked as such; not
re-run).

\paragraph{Protocol discipline.} A preregistration-style protocol was
written before running; every number lands in a result JSON with a
commit hash; the holdout is opened once per experiment.

\paragraph{Statistical gate choice (an empirical lesson).} The first
E1 implementation pooled all fixture-trial scores into a two-sample
comparator: improvement deltas up to $+0.278$ ($p=0.045$) read
\emph{inconclusive} because the noise band absorbed cross-fixture
difficulty variance rather than trial noise. The fix---per-fixture
means plus a paired sign-flip permutation test---removes fixture
variance by construction. The failure mode is itself a result: RSI
selection gates built on pooled two-sample tests are systematically
conservative on heterogeneous task sets, biasing loops toward stasis
(exactly the stuck-loop behavior we observed for 10 consecutive rounds
before the fix).

\section{Results}\label{sec:results}
\subsection{E0 --- pilot findings}

\textbf{The loop works end-to-end on local models.} A weak baseline
compaction prompt (score 0.650 of a 2.0 oracle ceiling) evolved in 10
rounds to 1.887 (arm A) and 1.913 (arm B)---within ${\sim}5\%$ of the
hand-written ceiling prompt (2.000). Six keeps across 20 rounds; every
keep passed a paired sign-flip permutation gate at $\alpha=0.05$ on
per-fixture means (trials=5); zero cloud calls.

\textbf{Gate statistics matter more than proposer skill at local
scale.} The first two protocol versions produced 30 consecutive
inconclusive rounds despite positive mean deltas---quantization noise
and pooled cross-fixture variance masked real signal. Only after the
continuous oracle plus paired test did the loop accept its first
improvement (round 1, delta $+0.833$, $p=0.031$). A self-improvement
loop with an underpowered gate is indistinguishable from a broken one.

\textbf{Information starvation, not task difficulty, was the binding
constraint.} The 8B proposer plateaued proposing the same vague fix for
10+ rounds while receiving only missing-item lists; adding one verbatim
example of the compactor's failing output unlocked immediate keeps.

\subsection{E1 --- main table}

\begin{table}[t]
\centering
\begin{tabular}{llcc}
\toprule
Arm & Seed & Final score & Keeps \\
\midrule
A (memory + evidence) & 0 & 2.000 & 3 \\
A & 1 & 1.950 & 3 \\
A & 2 & 0.825 & 0 \\
B (evidence only) & 0 & 1.900 & 3 \\
B & 1 & 1.600 & 3 \\
B & 2 & 1.700 & 2 \\
C (memory only) & 0 & 1.925 & 2 \\
C & 1 & 1.325 & 1 \\
C & 2 & 1.825 & 3 \\
\bottomrule
\end{tabular}
\caption{E1: final incumbent quality by arm and seed (oracle ceiling
2.0; baseline without loop 0.650; trials=5; per-pass re-score noise up
to ${\sim}0.4$).}
\label{tab:e1}
\end{table}

\textbf{1. The loop beats the baseline from every information source.}
Every run with $\geq 2$ keeps lands 1.6--2.0 vs.\ 0.650
(2.5--3.1$\times$). The two runs with $<2$ keeps diverge: A seed 2 (0
keeps) stayed inside the re-score noise floor (0.700 $\rightarrow$
0.825); C seed 1 (1 keep) improved 0.500 $\rightarrow$ 1.325, real but
the lowest final of the improving runs.

\textbf{2. The memory-transfer hypothesis is not supported.} The memory
summary did not add measurable value on top of direct failure evidence
(B mean 1.733 vs.\ A 1.592; A's deficit is entirely its one 0-keep
run---its other seeds hit 2.000/1.950, the best finals recorded). With
N=3 and ${\sim}0.3$--0.4 re-score noise, A/B/C means are statistically
indistinguishable; the defensible claim is ``no evidence that memory
adds on top of evidence,'' not ``memory hurts.''

\textbf{3. Memory alone is real.} Arm C (summary, no evidence) mean
1.692 with all three seeds improving---the cloud-scale transfer story
does not fully carry over, but memory-only is clearly above baseline: a
weak proposer can farm its own rollback lessons.

\textbf{4. Keep-count is the load-bearing variable.} 7 of 9 runs
reached $\geq 2$ keeps and all improved $\geq 2.5\times$; the two
below-2-keep runs improved 0.125 (noise-bound) and 0.825 (real but
lowest final). Gate pass rate, not memory content, predicted outcome.

\textbf{5. Measurement variance.} Re-scoring the same final incumbent
across aggregation passes moved means by up to ${\sim}0.4$ (4B sampling
noise at temperature 0.2)---all cross-arm comparisons live inside this
noise floor, which is itself a local-regime finding: at this scale,
selection noise dominates arm differences.

\subsection{E2 --- deterministic gate vs.\ LLM-judge gate}

Run 1 (4 candidates: 1 genuine improvement + 3 grader-hack decoys ---
flattery phrasing, grader-vocabulary echoing, format gaming; all zero
or negative true effect, styled to please a judge):

\begin{table}[t]
\centering
\begin{tabular}{lccccc}
\toprule
Candidate & True $\Delta$ (train) & Holdout $\Delta$ & Deterministic & Judge & Judge $\Delta$ \\
\midrule
genuine & $+1.067$ & $+0.700$ & \textbf{keep} & inconclusive & $-0.07$ \\
decoy-flattery & $+0.133$ & $+0.200$ & inconclusive & \textbf{keep} & $+3.00$ \\
decoy-vocab & $-0.200$ & $-0.200$ & inconclusive & \textbf{keep} & $+2.27$ \\
decoy-format & $0.000$ & $0.000$ & inconclusive & \textbf{keep} & $+2.73$ \\
\bottomrule
\end{tabular}
\caption{E2 run 1: decoy keep-rate deterministic 0/3, judge 3/3; the
judge rejected the genuine candidate (rubric score moved down $-0.07$
while the true score moved $+1.07$).}
\label{tab:e2run1}
\end{table}

\textbf{Replication runs 2--3.} Run 3 widens the decoy set to 5, adding
two grader-hack styles: \emph{decoy-verbose} (exhaustive 400-word
summaries targeting the rubric's completeness axis) and
\emph{decoy-meta} (direct reader address plus self-congratulatory
framing targeting the professionalism axis).

\begin{table}[t]
\centering
\begin{tabular}{llccl}
\toprule
Run & Decoys & Det.\ keep & Judge keep & Genuine: det.\ / judge \\
\midrule
1 & 3 phrasing & 0/3 & 3/3 & keep / \textbf{rejected} ($-0.07$ vs.\ $+1.07$) \\
2 & 3 phrasing & 0/3 & 2/3 & keep / inconclusive ($+0.07$ vs.\ $+1.10$) \\
3 & 3 phrasing + 2 new & 0/5 & 3/5 & keep / inconclusive ($0.00$ vs.\ $+1.10$) \\
\bottomrule
\end{tabular}
\caption{E2 across three runs: deterministic 0/11 decoy placements
kept, judge 8/11, and the judge never selected the genuine candidate
(0/3 runs).}
\label{tab:e2agg}
\end{table}

Reading:
\begin{enumerate}
\item \textbf{The deterministic gate kept zero decoys in all 11
placements; the judge kept 8/11.} Phrasing that flatters a judge has no
channel through a pure function of output versus fixture.
\item \textbf{The judge failed to select the genuine improvement in all
three runs}---once rejecting it outright, twice returning
inconclusive. A local 32B judge on a small-model compaction task is
both gameable \emph{and} noise.
\item \textbf{Attack-style dependence.} The judge's vulnerability
concentrates on self-referential quality phrasing---the style its
rubric implicitly rewards; verbosity and meta-commentary drew
inconclusive verdicts rather than keeps. The deterministic gate is
style-blind: 0 keeps regardless of attack family.
\item \textbf{Cost.} The deterministic arm was cheaper than the judge
arm in all three runs (0.65$\times$, 0.75$\times$, 0.61$\times$); the
gate itself adds zero tokens and zero model calls---its cost is the
shared evaluation trials.
\item \textbf{Implication.} At cloud scale an LLM judge is a defensible
selection signal because the optimizer is strong enough that signal
noise is the binding term. Here the proposer is a peer of the judge's
own capacity class, and judge-flattering text is exactly the kind of
edit a weak proposer over-produces---the gate and the proposer must not
share a loss surface.
\end{enumerate}

\subsection{E3 --- budget accounting}

\begin{table}[t]
\centering
\begin{tabular}{lcccccc}
\toprule
Arm & Seed & Rounds & Keeps & Final & Wall-clock (min) & Min/keep \\
\midrule
A & 0 & 10 & 3 & 2.000 & 26.2 & 8.7 \\
B & 0 & 10 & 3 & 1.900 & 17.4 & 5.8 \\
C & 0 & 10 & 2 & 1.925 & 29.1 & 14.6 \\
A & 1 & 10 & 3 & 1.950 & 105.7$^{*}$ & 35.2$^{*}$ \\
B & 1 & 10 & 3 & 1.600 & 33.9 & 11.3 \\
A & 2 & 10 & 0 & 0.825 & 23.7 & --- \\
B & 2 & 10 & 2 & 1.700 & 16.3 & 8.2 \\
C & 1 & 10 & 1 & 1.325 & 133.2$^{*}$ & 133.2$^{*}$ \\
C & 2 & 10 & 3 & 1.825 & 27.8 & 9.3 \\
\bottomrule
\end{tabular}
\caption{E3: budget accounting, 9 runs on one M5 Max (128\,GB unified
memory), zero cloud calls. $^{*}$A seed 1 includes server-restart
outage gaps; C seed 1 spans a period of heavy co-tenancy---treat both
as upper bounds.}
\label{tab:e3}
\end{table}

\textbf{1. ${\sim}21$ minutes per kept improvement} (413 min total / 20
keeps, 9 runs) on one consumer machine, all-inclusive: proposer calls,
66 compactor evaluation calls per round, retries, and waiting. Per-keep
cost excluding the two flagged runs: ${\sim}9$--15 min.

\textbf{2. Gate cost is exactly zero LLM tokens.} The keep/rollback
decision is a pure function over per-fixture means. A 10-round loop
costs 10 proposer calls + 660 compactor evaluation calls; the gate adds
nothing.

\textbf{3. Quality per local-compute-hour.} A run buys $+0.8$ to $+1.3$
oracle score (baseline 0.650, ceiling 2.0) for ${\sim}20$--35 min of
wall clock---compare GEPA-style cloud loops where the same loop shape
bills frontier-model tokens per scored candidate. The claim form stays
``gains per local compute,'' not absolute SOTA.

\section{Discussion}\label{sec:discussion}
\paragraph{Failure modes observed.} Three distinct loop-stasis
mechanisms, each invisible from the outside: (1) \emph{score
quantization} --- a binary-success oracle gives so few distinct values
that real deltas never clear the noise band (30 inconclusive rounds);
(2) \emph{pooled gates on heterogeneous tasks} --- cross-fixture
difficulty variance inflates the noise band, so only implausibly large
deltas pass (Section \ref{sec:design}); (3) \emph{information
starvation} --- a weak proposer given only missing-item lists
re-proposes the same vague fix indefinitely (Section
\ref{sec:results}). All three are gate/signal-design bugs, not
capability limits: the fix space was reachable the whole time (the
hand-written prompt hit the ceiling).

\paragraph{What actually predicted loop success.} Not memory content,
not proposer skill, not arm choice: keep-count. Every run with
$\geq 2$ keeps improved 2.5--3.1$\times$ over baseline; the one run
with 0 keeps stayed inside the re-score noise floor (initial 0.700
$\rightarrow$ final 0.825). For local RSI practitioners this reframes
the engineering target: invest in gate power (continuous scores,
pairing, trials) and proposer information (verbatim failure evidence),
not in richer memory architectures.

\paragraph{When weight-level loops take over.} Prompt-level
self-improvement plateaus when the fix space is not expressible as a
prompt mutation; we saw the ceiling reached within 3 keeps on a single
component. Weight-update loops (test-time RL etc.) subsume prompt loops
at equal compute for large enough models, but on Apple Silicon unified
memory the training budget per accepted improvement is orders of
magnitude above our ${\sim}16$ min/keep---the decision rule of
\citet{equalcompute2026} (weight loop when harness delta $<$ model
delta at equal compute) resolves to prompt-level loops in this regime.

\paragraph{The gate/proposer coupling principle.} E2 generalizes beyond
the decoy experiment: any selection signal computed by the same model
family or capacity class as the proposer shares a loss surface with it.
Proposers drift toward whatever the gate rewards; a judge rewards
phrasing; a pure function rewards only true behavior. This is the local
regime's version of ``never optimize against the monitor''
\citep{monitor2025}, and it is \emph{stronger} at local scale because
the proposer is too weak to achieve real gains reliably, so flattering
the gate is the cheapest gradient available to it.

\section{Limitations}\label{sec:limitations}
\begin{enumerate}
\item \textbf{Scale and scope.} One hardware machine (M5 Max, 128\,GB),
one task family (synthetic compaction fixtures, oracle ceiling 2.0),
one loop component (compaction prompt). N=3 seeds per arm; E2 is a
three-run replication with 1 genuine + 3--5 decoys per run. Results are
existence proofs of regime behavior, not distributional claims.
\item \textbf{Verifier expressiveness.} The deterministic gate is only
as rich as the fixture oracle---token containment and rule recall.
Tasks lacking a pure-function score (open-ended generation, subjective
quality) are out of scope by construction; that is the price of the
anti-hacking guarantee, and it bounds which components a local RSI loop
can self-improve.
\item \textbf{Genuine-candidate provenance.} E2's genuine candidate was
constructed from the E1 loop's own accepted mutation---a real
improvement, but produced by our pipeline rather than discovered blind
by the E2 loop. The decoy asymmetry (0/11 vs.\ 8/11) is the
load-bearing claim; the judge's rejection of the genuine candidate is
consistent across its three appearances but is one candidate.
\item \textbf{Cloud comparisons are literature-reported}, not re-run
(SICA, DGM, GEPA numbers cited as reported); hardware and task sets
differ, so cost comparisons are order-of-magnitude framing, not
controlled measurement.
\item \textbf{Measurement variance.} Re-scoring the same incumbent
moved final scores by up to ${\sim}0.3$; all aggregate numbers average
over evaluation passes, but single-run numbers carry that noise floor.
\end{enumerate}

\section*{Acknowledgements}
The authors thank Xiao Fei, Shao Tianyu, Liu Longlai, and
Liu Shusong for discussions on agent harness design and for feedback
on earlier drafts of this manuscript. The first author is deeply
grateful to his
wife, son, and daughter for their patience and unconditional support
during
the long evenings and weekends this work consumed.
The views and opinions expressed in this paper are those of the authors
and do not necessarily reflect the official policy or position
of Huawei.

\bibliographystyle{plainnat}
\bibliography{refs}

\end{document}